\def\year{2020}\relax
\documentclass[letterpaper]{article} 
\usepackage{aaai20}  
\usepackage{times}  
\usepackage{helvet} 
\usepackage{courier}  
\usepackage[hyphens]{url}  
\usepackage{graphicx} 
\usepackage{graphicx}  
\usepackage{amsmath,amsfonts,bm}
\usepackage{booktabs}
\usepackage{multirow}
\newcommand{\eqform}[1]{Equation~(\ref{#1})}
\newcommand{\sptk}[1]{\texttt{[#1]}}
\newcommand{\Ls}{\mathcal{L}}
\newcommand{\train}{\mathcal{D}}

\newcommand{\namecite}[1]{\citeauthor{#1}~\shortcite{#1}}

\newcommand\ours{\textsc{Xnlg}}
\newcommand\xlm{\textsc{Xlm}}
\newcommand\txlmt{\textsc{Pipeline}~(\xlm)}
\newcommand\xlmt{\textsc{Pipeline}~(\xlm)}
\newcommand\txlm{\textsc{Pipeline}~(\xlm)}
\newcommand\ggtpp{~~~ w/ Google Translator}

\def\trainmono{{\train_{\textnormal{m}}}}
\def\trainpara{{\train_{\textnormal{p}}}}

\title{Cross-Lingual Natural Language Generation via Pre-Training}
\author{Zewen Chi{$^\dag$}\thanks{\ \  Contribution during internship at Microsoft Research.},~~Li Dong\textsuperscript{$\ddagger$},~~Furu Wei\textsuperscript{$\ddagger$},~~Wenhui Wang\textsuperscript{$\ddagger$},~~Xian-Ling Mao\textsuperscript{$\dag$},~~Heyan Huang\textsuperscript{$\dag$} \\
\textsuperscript{$\dag$}Beijing Institute of Technology \\
\textsuperscript{$\ddagger$}Microsoft Research\\
\texttt{\{czw,maoxl,hhy63\}@bit.edu.cn}\\
\texttt{\{lidong1,fuwei,Wenhui.Wang\}@microsoft.com}}

\begin{document}

\maketitle

\begin{abstract}
In this work we focus on transferring supervision signals of natural language generation (NLG) tasks between multiple languages. We propose to pretrain the encoder and the decoder of a sequence-to-sequence model under both monolingual and cross-lingual settings. The pre-training objective encourages the model to represent different languages in the shared space, so that we can conduct zero-shot cross-lingual transfer. After the pre-training procedure, we use monolingual data to fine-tune the pre-trained model on downstream NLG tasks. Then the sequence-to-sequence model trained in a single language can be directly evaluated beyond that language (i.e., accepting multi-lingual input and producing multi-lingual output). Experimental results on question generation and abstractive summarization show that our model outperforms the machine-translation-based pipeline methods for zero-shot cross-lingual generation. Moreover, cross-lingual transfer improves NLG performance of low-resource languages by leveraging rich-resource language data. Our implementation and data are available at~\url{https://github.com/CZWin32768/xnlg}.
\end{abstract}

\section{Introduction}

Learning natural language generation (NLG) models heavily relies on annotated training data.
However, most available datasets are collected in a single language (typically English), which restricts deploying the applications to other languages.
In this work, we aim at transferring the supervision of a monolingual NLG dataset to unseen languages, so that we can boost performance for the low-resource settings.


\begin{figure}[t]
\begin{center} 
\includegraphics[width=0.96\linewidth]{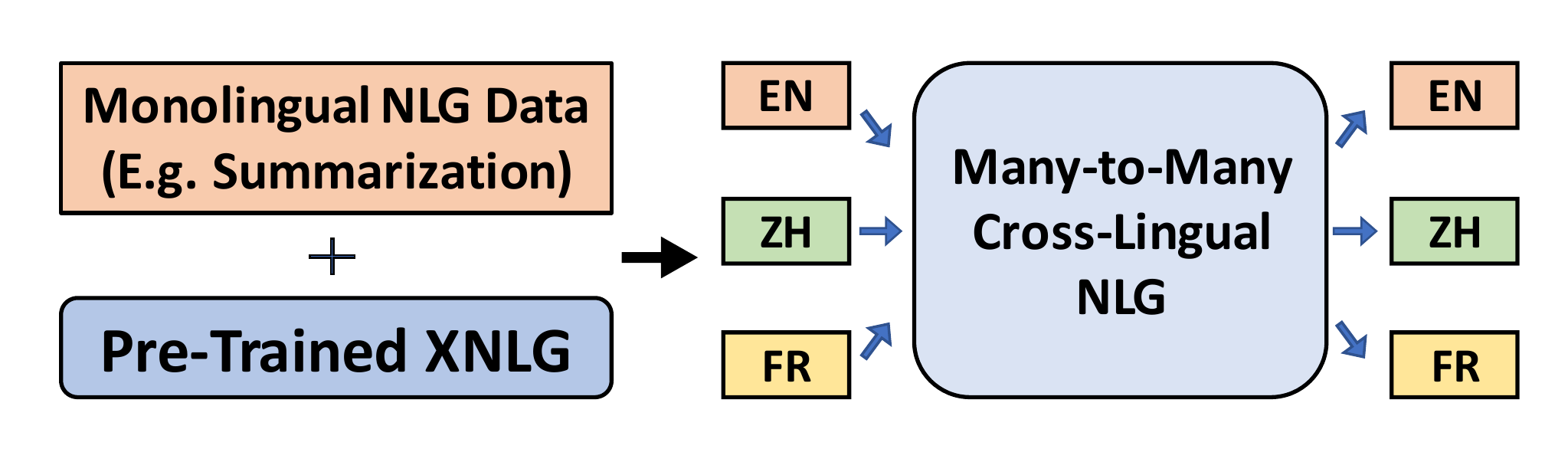}
\caption{We use a monolingual (such as English) NLG dataset to fine-tune the pre-trained model \ours{}, and then evaluate it beyond the language for both source and target sides (e.g., Chinese, and French).} 
\label{fig:intro}
\end{center}
\end{figure}

Various methods have been proposed over the years to learn cross-lingual word embeddings~\cite{mikolov2013exploiting,xing2015normalized,conneau2017word} or sentence encoders~\cite{johnson2017google,xnli,xlm}, which try to encode multilingual texts into a shared vector space.
Despite achieving promising results on cross-lingual classification problems, cross-lingual pre-trained models purposed for NLG tasks remains relatively understudied.

The cross-lingual generation problem is challenging due to the following reasons.
First, it requires the models to understand multilingual input texts, and generate multilingual target sequences. So both the encoder and the decoder should be pre-trained together.
Second, the many-to-many nature of cross-lingual NLG increases language pairs with the square of the number of languages.
Third, the prediction space of cross-lingual NLG is much larger than classification tasks, which makes knowledge transfer of decoders quite critical.

Previous work mainly relies on machine translation (MT) to map texts to different languages.
The first strand of research directly uses MT in a pipeline manner~\cite{pipeline2010wan}.
For example, the inputs written in other languages are first translated to English, and fed into the NLG model that is trained by English data. Then the generated English texts are translated back to the target language.
Another strand of work uses MT to generate pseudo training data for other language pairs that are lack of annotations~\cite{xnhg,xsummacl}.
However, such methods have to use multiple MT systems, which renders them suffering from error propagation.
Moreover, because the pipeline-based methods do not explicitly share the same parameter space across languages, we can not directly transfer the task-specific supervision to other low-resource languages.

In this paper, we propose a cross-lingual pre-trained model (named as \ours{}) in order to transfer monolingual NLG supervision to other pre-trained languages by fine-tuning.
Specifically, \ours{} shares the same sequence-to-sequence model across languages, and is pre-trained with both monolingual and cross-lingual objectives.
The model not only learns to understand multilingual input, but also is able to generate specific languages by conditioning on the encoded semantics.
Figure~\ref{fig:intro} demonstrates how to use \ours{} to perform cross-lingual transfer for downstream tasks.
The proposed model enables us to fine-tune the pre-trained model on monolingual NLG training data, and then evaluate it beyond a single language, including zero-shot cross-lingual generation.
Besides, we explore several fine-tuning strategies to make a compromise between cross-lingual ability and task ability.
In addition, we introduce two cross-lingual NLG datasets (i.e., question generation, and abstractive summarization) for evaluation, which includes three languages, namely English, Chinese, and French.
Experimental results on the NLG tasks show that \ours{} achieves competitive performance compared with the machine-translation-based pipeline model in zero-shot cross-lingual settings.

\section{Related Work}

\paragraph{Cross-Lingual NLG}

Several previous methods have been proposed for cross-lingual abstractive summarization.
\namecite{xnhg} and \namecite{xsummacl} use translated documents or summaries as pseudo training data.
\namecite{ncls} incorporate monolingual summarization and machine translation to improve cross-lingual summarization.
However, the systems only conduct experiments that generate summaries with different languages from the input language, rather than transferring supervision signals across all language pairs.
\namecite{kumar2019cross} use training data annotated in multiple languages to jointly train a sequence-to-sequence model for question generation.
In contrast, our method can also be applied to zero-shot settings across languages.

\paragraph{Monolingual Pre-Training}


Various training objectives are designed to pretrain text encoders used for general-purpose representations, such as language modeling~\cite{elmo,gpt,bert,spanbert,xlnet}, auto-encoding~\cite{pretrain-dae}, and machine translation~\cite{mccann2017learned}.
Apart from pre-training encoders, several pre-trained models~\cite{unilm,mass} are proposed for generation tasks.
In comparison, our goal is to investigate a pre-training method for cross-lingual NLG tasks.

\paragraph{Cross-Lingual Pre-Training}


By pre-training BERT~\cite{bert} on corpus of multiple languages, it shows a surprising ability to produce cross-lingual representations~\cite{wu2019beto}.
More recently, \namecite{xlm} extend mask language modeling pre-training to cross-lingual settings, which shows significant improvements on cross-lingual classification and unsupervised machine translation.
By comparison, we pretrain both encoder and decoder for cross-lingual generation tasks, rather than only focusing on encoder.
\namecite{artetxe2018massively} use the sequence encoder of the multilingual translation model~\cite{johnson2017google} to produce cross-lingual sentence embeddings.
However, as shown in the experiments (Section~\ref{sec:exp}), it is difficult to control the target language by directly fine-tuning the pre-trained translation model on downstream NLG tasks.

\section{Methods}




\begin{figure*}[th]
\begin{center} 
\includegraphics[width=0.98\linewidth]{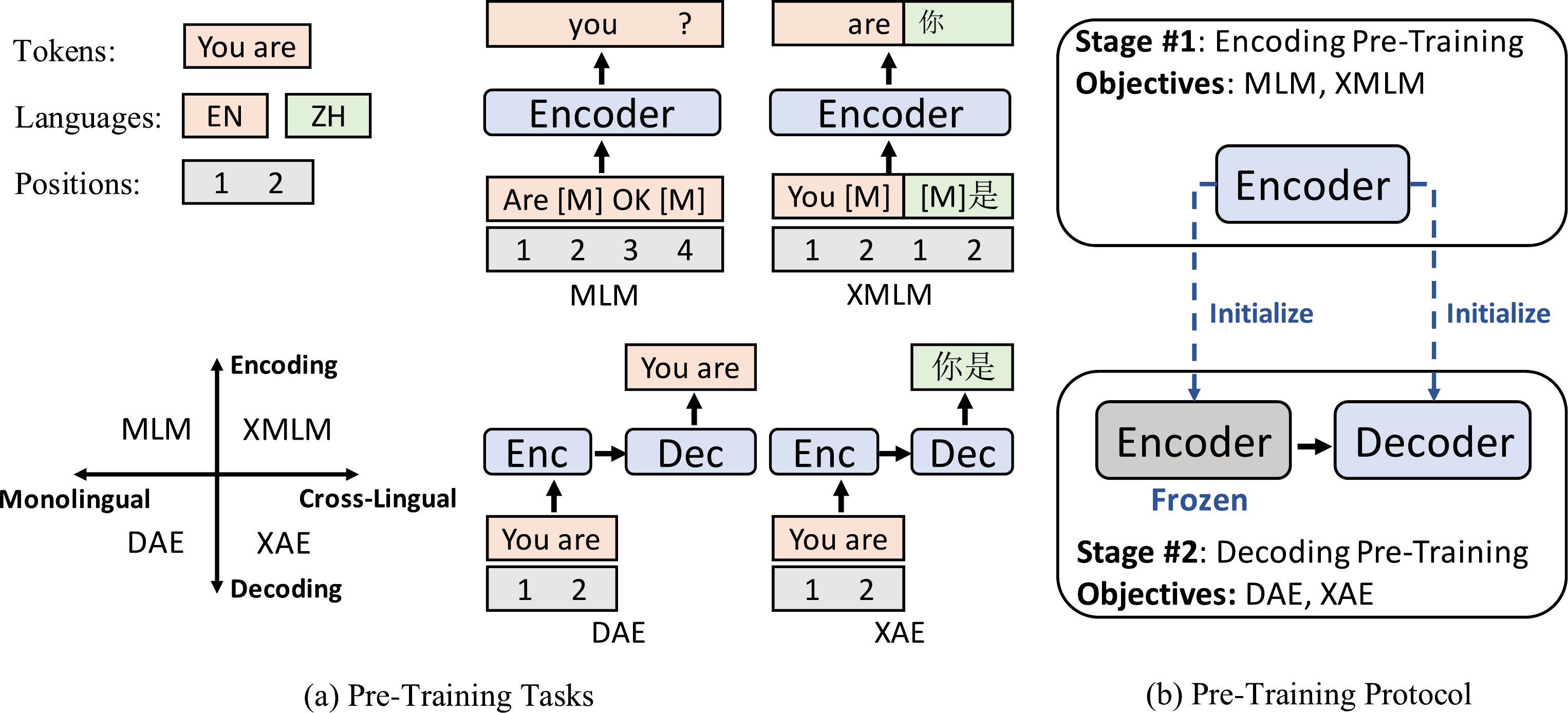}
\caption{Overview of the pre-training tasks and the pre-training protocol designed for \ours{}.
}
\label{fig:ov}
\end{center} 
\end{figure*}

As shown in Figure~\ref{fig:ov}, \ours{} is a pre-trained sequence-to-sequence model, which is based on Transformer~\cite{transformer}.
Both the encoder and the decoder are supposed to support multiple languages.
Following~\cite{xlm}, we use language tag embeddings to distinguish the source and target languages.
Given a sentence and its corresponding language tag, \ours{} encodes the input into vector representations.
By conditioning on the encoding vectors and a specific language tag, the decoder generates the output sequence in the target language.

\subsection{Pre-Training Tasks}

\paragraph{Monolingual MLM}

The masked language modeling (MLM)~\cite{bert} task aims at predicting the randomly masked words according to their context.
The objective pretrains the bidirectional encoder to obtain contextual representations.
Following~\cite{bert}, we randomly mask 15\% of the tokens in a monolingual sentence.
For each masked token, we substitute it with a special token \sptk{M}, a random token, or the unchanged token with probabilities of 0.8, 0.1, and 0.1, respectively.
Let $x$ denote a sentence from the monolingual training corpus, and $M_{x}$ the set of randomly masked positions.
The monolingual MLM loss is defined as:
\begin{align}
\Ls_{\textnormal{MLM}}^{(x)} = -\sum_{i \in M_{x}}\log p( x_i | x_{\setminus M_{x}}) \label{eq:mlm}
\end{align}
where $x_{\setminus M_{x}}$ is the masked version of input $x$.
The language tags are fed into the model for all pre-training tasks.

\paragraph{Denoising Auto-Encoding (DAE)}

We use the denoising auto-encoding (DAE) objective~\cite{dae} to pretrain the encoder-decoder attention mechanism.
Given sentence $x$ from the monolingual corpus, we use three types of noise to obtain the randomly perturbed text $\hat{x}$.
First, the word order is locally shuffled.
Second, we randomly drop tokens of the sentence with a probability of $0.1$.
Third, we substitute tokens with the special padding token \sptk{P} with a probability of $0.1$.
The pre-training objective is to recover the original sentence $x$ by conditioning on $\hat{x}$.
The DAE loss is computed via:
\begin{align}
\Ls_{\textnormal{DAE}}^{(x)} = -\log p(x|\hat{x}) = -\sum_{i = 1}^{|x|}{\log p(x_i | \hat{x}, x_{<i})}
\end{align}
where $x_{<i} = x_1,\cdots,x_{i-1}$.

\paragraph{Cross-Lingual MLM (XMLM)}
Similar to monolingual MLM, the masked token prediction task can be extended to cross-lingual settings~\cite{xlm}.
To be specific, given a parallel corpus, we concatenate the pair of bilingual sentences $(x,y)$ to a whole sequence, and use it as the input of MLM.
The language tags are also fed into the model to indicate the languages of tokens.
During training, we adopt the same masking strategy as monolingual MLM.
Apart from using monolingual context to predict the masked tokens, XMLM encourages the model to utilize the alignment of bilingual sentences, so that the model learns to map cross-lingual texts into a shared vector space.
Similar to \eqform{eq:mlm}, the cross-lingual MLM loss is:
\begin{align}
\Ls_{\textnormal{XMLM}}^{(x,y)} = -\sum_{i \in M_x}{\log p( x_i | x_{\setminus M_x} , y_{\setminus M_y})} \\
-\sum_{i \in M_y}{\log p( y_i | x_{\setminus M_x} , y_{\setminus M_y})}
\end{align}
where $M_x, M_y$ represent the masked positions of $x$ and $y$, respectively.

\paragraph{Cross-Lingual Auto-Encoding (XAE)}

If only DAE is used as the pre-training task for the decoder, we found that the model ignores the target language tag while generating just the same language as the input, caused by the spurious correlation issue~\cite{spuriouscorr}.
In other words, the DAE loss captures the spurious correlation between the source language tag and the target sentences, but we expect the language of generated sentences can be controlled by the target language tag.
To solve the above problem, we use machine translation as the cross-lingual auto-encoding (XAE) task, which decreases mutual information between the target sentences and the source language tag.
XAE can be viewed as the multilingual-version DAE task in the sense that both of them recover the sentence by conditioning on the encoded representations.
The cross-lingual auto-encoding loss is:
\begin{align}
\Ls_{\textnormal{XAE}}^{(x,y)} = -\log p(y|x) - \log p(x|y)
\end{align}
where $(x,y)$ is a pair of sentences in the parallel corpus.




\subsection{Pre-Training Protocol}

As shown in Figure~\ref{fig:ov}(b), we propose a two-stage pre-training protocol for \ours{}.
The first stage pretrains the encoding components, where the model learns to encode multilingual sentences to a shared embedding space.
We consider using MLM and XMLM as the pre-training tasks.
The objective of the first stage is to minimize:
\begin{align}
\Ls_{1}= \sum_{(x,y) \in \trainpara} \Ls_{\textnormal{XMLM}}^{(x,y)} + \sum_{x \in \trainmono} \Ls_{\textnormal{MLM}}^{(x)}
\end{align}
where $\trainpara$ indicates the parallel corpus, and $\trainmono$ is the monolingual corpus.

Although the pre-trained encoder in the first stage enables the model to encode multilingual sentences. However, it cannot directly be used in cross-lingual NLG because: 1) encoder-decoder attention is not pre-trained; 2) the decoding algorithm is different between masked language modeling and autoregressive decoding, resulting in the mismatch between pre-training and fine-tuning.
Therefore, we conduct decoding pre-training in the second stage by using DAE and XAE as the tasks.
Besides, we only update decoder parameters and keep the encoder fixed.
The objective of the second stage is to minimize:
\begin{align}
\Ls_{2} = \sum_{(x,y) \in \trainpara}{\Ls_{\textnormal{XAE}}^{(x,y)}} + \sum_{x \in \trainmono}{\Ls_{\textnormal{DAE}}^{(x)}}
\end{align}

\subsection{Fine-Tuning on Downstream NLG Tasks}

In the fine-tuning procedure, let us assume that we only have English training data for downstream NLG tasks.
According to whether the target language is English, the directions of NLG can be categorized into two classes: any languages to non-English languages (Any-to-Others), and any languages to English (Any-to-English).

\paragraph{Fine-Tuning for Any-to-Others NLG}

Ideally, the model can be fine-tuned towards a new task without losing its cross-lingual ability.
However, we observe the catastrophic forgetting of target language controllability, if we fine-tune all the model parameters for Any-to-Others NLG.
So we keep the decoder and the word embeddings frozen and only update the encoder parameters during fine-tuning.
In practice, we found that the proposed fine-tuning method prevents the model from only decoding English words for the Any-to-Others setting.

\paragraph{Fine-Tuning for Any-to-English NLG}

For the Any-to-English NLG transfer, the decoder always generates English. So we can freeze the encoder parameters, and update the decoder parameters to retain the cross-lingual ability. 
As an alternative way, we can also fine-tune all the parameters to obtain the best results on the English dataset while having a slight drop in performance.


\begin{table}[t]
\centering
\small
\begin{tabular}{llccc}
\toprule
Data                   & Models             & BL-4 & MTR & RG-L \\ \midrule
\multirow{4}{*}{En-QG} & \textsc{CorefNqg}$\dag$    & 15.16  & 19.12  & -       \\
                       & \textsc{Mp-Gsn}      & 16.38  & 20.25  & 44.48   \\
                       & \xlm{}               & 16.94  & 21.87  & 46.45   \\
                       & \ours{}              & \textbf{19.99} & \textbf{24.05} & \textbf{48.74} \\ \midrule
\multirow{2}{*}{Zh-QG} & \xlm{}              & 23.41     & 23.32     & 47.40     \\
                       & \ours               & \textbf{24.89} & \textbf{24.53} & \textbf{49.72}     \\

\bottomrule
\end{tabular}
\caption{Evaluation results of monolingual supervised question generation for English and Chinese. BL is short for BLEU, MTR for METEOR, and RG for ROUGE. The results with ``$\dag$'' are reported on different data splits.}
\label{table:en-en-qg}
\end{table}

\begin{table}[t]
\centering
\small
\begin{tabular}{lccc}
\toprule
 Models  & BL-4    & MTR    & RG-L   \\ \midrule
 \xlm{}   & 0.25      & 0.62      & 2.56      \\
 \txlmt{}  & 4.42      & 9.59      & 21.22     \\
 \ggtpp{} & 9.95      & 14.92     & 29.37     \\
 \ours    & \textbf{16.37}     & \textbf{18.74}     & \textbf{34.93}      \\ \bottomrule
\end{tabular}
\caption{Evaluation results of zero-shot Chinese-Chinese question generation. Same shorthands apply as in Table~\ref{table:en-en-qg}.}
\label{table:zh-zh-qg}
\end{table}

\section{Experiments}
\label{sec:exp}

We conduct experiments over two cross-lingual NLG downstream tasks, i.e., cross-lingual question generation, and cross-lingual abstractive summarization.
We compare \ours{} with state-of-the-art cross-lingual pre-trained models, and machine-translation-based pipelines.

\begin{table}[t]
\centering
\small
\begin{tabular}{lccc}
\toprule
Models  & Rel & Flu & Corr   \\ \midrule
\xlm{}     & 0~~        & 0~~       & 0~~      \\
\txlmt{}    & 0.50~~     & 0.80~~    & 0.03~~       \\
\ggtpp{}    & 1.31~~    & \textbf{1.43}*    & 0.69~~    \\
\ours      & \textbf{1.68}*     & 1.29~~    & \textbf{0.89}*      \\
\bottomrule
\end{tabular}
\caption{Human evaluation results of zero-shot Chinese-Chinese question generation. Rel is short for relatedness, Flu for fluency, and Corr for correctness. ``*'' indicates the improvements are significant at $p < 0.05$. }
\label{table:human-zh}
\end{table}

\begin{table}[t]
\centering
\small
\begin{tabular}{lccc}
\toprule
Models  & Rel & Flu & Corr   \\ \midrule
\xlm{}     & 0~~        & 0~~       & 0~~      \\
\xlmt{}    & 0.87~~     & 0.86~~    & 0.28~~       \\
\ours      & \textbf{1.09}*     & \textbf{0.95}~~    & \textbf{0.53}*      \\ \bottomrule
\end{tabular}
\caption{Human evaluation results of zero-shot English-Chinese question generation. ``*'' indicates the improvements are significant at $p < 0.05$. Same shorthands apply as in Table \ref{table:human-zh}.}
\label{table:en-zh-qg}
\end{table}

\begin{table}[t]
\centering
\small
\begin{tabular}{lccc}
\toprule
Models  & Rel & Flu & Corr   \\ \midrule
\xlm{}   & 1.00~~  & 1.20~~ & 0.40~~  \\
\txlm{}  & 0.85~~  & 0.98~~ & 0.28~~  \\
\ours    & \textbf{1.24}*  & \textbf{1.47}* & \textbf{0.76}*  \\ \bottomrule
\end{tabular}
\caption{Human evaluation results of zero-shot Chinese-English question generation. ``*'': the improvements are significant at $p < 0.05$. Same shorthands apply as in Table \ref{table:human-zh}.}
\label{table:zh-en-qg}
\end{table}

\begin{table}[t]
\centering
\small
\begin{tabular}{llccc}
\toprule
Data                   & Models   & RG-1      & RG-2      & RG-L   \\ \midrule
\multirow{2}{*}{En-AS} & \xlm{}   & 48.15     & 26.35     & 45.04    \\
                       & \ours    & \textbf{48.76}     & \textbf{26.82}     & \textbf{45.57}     \\ \midrule
\multirow{2}{*}{Fr-AS} & \xlm{}   & 56.27     & 39.20     & 52.84     \\
                       & \ours    & \textbf{57.84}     & \textbf{40.81}     & \textbf{54.24}     \\ \midrule
\multirow{2}{*}{Zh-AS} & \xlm{}   & 55.30     & 42.57     & 52.95     \\
                       & \ours    & \textbf{57.65}     & \textbf{44.93}     & \textbf{54.95}     \\
\bottomrule
\end{tabular}
\caption{Evaluation results of supervised monolingual summarization. Same shorthands apply as in Table~\ref{table:en-en-qg}.}
\label{table:mono-as}
\end{table}

\begin{table}[t]
\centering
\small
\begin{tabular}{llccc}
\toprule
Models     & RG-1      & RG-2      & RG-L   \\ \midrule
\xlm{}   & 14.53     & 1.80      & 13.43     \\
\txlmt{} & 30.58     & 12.01     & 27.44          \\
\ggtpp{} & 38.48     & 18.86     & 34.98      \\
\ours    & \textbf{39.98}     & \textbf{20.31}     & \textbf{36.31}     \\ \bottomrule
\end{tabular}
\caption{Evaluation results of zero-shot French abstractive summarization. Same shorthands apply as in Table~\ref{table:en-en-qg}.}
\label{table:fr-fr-as}
\end{table}

\begin{table}[t]
\centering
\small
\begin{tabular}{lccc}
\toprule
Models     & RG-1      & RG-2      & RG-L   \\ \midrule
\xlm{}     & 0.71      & 0.28      & 0.70      \\
\txlmt{}   & 26.39     & 13.11     & 23.98     \\
\ggtpp{}   & 36.96     & 22.03     & 33.99     \\
\ours      & \textbf{41.66}     & \textbf{28.70}     & \textbf{38.91}     \\ \bottomrule
\end{tabular}
\caption{Evaluation results of zero-shot Chinese abstractive summarization. Same shorthands apply as in Table~\ref{table:en-en-qg}.}
\label{table:zh-zh-as}
\end{table}

\subsection{Training Details}

\paragraph{Pre-Training}
We use a pre-trained \ours{} with a 10-layer encoder and a 6-layer decoder. For every Transformer layer, we use 1024 hidden units, 8 attention heads, and GELU activations~\cite{gelu}.
In the first pre-training stage, we directly use the 15-language pre-trained XLM~\cite{xlm} to initialize the parameters of our encoder and decoder. 
In the second stage, we use Wikipedia as the monolingual data for the DAE objective, and MultiUN~\cite{multiun} as the parallel data for the XAE objective.
The DAE loss is trained with a weight of $0.5$.
We train a two-language (English/Chinese) and a three-language (English/French/Chinese) \ours{} for two downstream NLG tasks, respectively.
Following~\cite{xlm}, we use the tokenizer provided by~\cite{chang2008optimizing} for Chinese, and Moses\footnote{\url{https://github.com/moses-smt/mosesdecoder}} for other languages, respectively. Then the words in all languages are split with a shared subword vocabulary learned by BPE~\cite{bpe}. We use Adam optimizer with a linear warm-up over the first 4,000 steps and linear decay for later steps, and the learning rate is set to $10^{-4}$.
The pre-training batch size is 64, and the sequence length is set to 256.
It takes about 30 hours to run 23,000 steps for the pre-training procedure by using 4 Nvidia Telsa V100-16GB GPUs.

\paragraph{Fine-Tuning}

For fine-tuning on downstream NLG tasks, we use Adam optimizer with a learning rate of $5\times10^{-6}$. We set the batch size as 16 and 32 for question generation and abstractive summarization, respectively. When the target language is the same as the language of training data, we fine-tune all parameters. When the target language is different from the language of training data, we fine-tune the Transformer layers of the encoder. We truncate the input sentences to the first 256 tokens. During decoding, we use beam search with a beam size of 3, and limit the length of the target sequence to 80 tokens.

\subsection{Question Generation}

We evaluate our model on zero-shot cross-lingual answer-aware question generation (QG). The goal is to generate a question that asks towards the answer with the given passage and the expected answer.
In the following experiments, we extend the QG task to the cross-lingual setting. By only using English QG training data, our goal is to generate questions in English or Chinese with the given passage-answer pair in English or Chinese.

We use SQuAD 1.1~\cite{squad1} as the English QG dataset.
It is a popular English question answering dataset containing over 100,000 questions and their corresponding annotated passages.
Following \cite{zhao-qg-2018}, we regard the original development set as the test set, and sample 5000 examples from the training data of two datasets as the development sets.
For Chinese QG, we follow the default data splits of WebQA~\cite{webqa}.
We regard the provided annotated evidence sentences as the input passages instead of entire documents.
To construct the input sequence, we view the whole input passage as a single sentence, and concatenate the passage and the answer into one sequence with a special token \sptk{S} between them.
During decoding Chinese, we utilize a subset of vocabulary, which is obtained from the passage sentences of the WebQA dataset.



\paragraph{English-English Question Generation}
We first conduct experiments on the supervised English-English QG setting. We compare our model to the following baselines:

\begin{itemize}
\item \textbf{\textsc{CorefNqg}}~\cite{du-qg-2018} An attentional sequence-to-sequence model with a feature-rich encoder.
\item \textbf{\textsc{Mp-Gsn}}~\cite{zhao-qg-2018} A sequence-to-sequence model with self-attention and maxout pointer mechanism.
\item \textbf{\xlm}~\cite{xlm} State-of-the-art cross-lingual pre-trained Transformer. We initialize the sequence-to-sequence model with pre-trained XLM.
\end{itemize}

We evaluate models with BLEU-4 (BL-4), ROUGE (RG) and METEOR (MTR) metrics.
As shown in Table~\ref{table:en-en-qg}, \ours{} outperforms the baselines, 
which demonstrates that our pre-trained model provides a good initialization for NLG.


\paragraph{Chinese-Chinese Question Generation}
We conduct experiments on zero-shot Chinese-Chinese QG to evaluate the cross-lingual transfer ability. In this task, models are trained with English QG data but evaluated with Chinese QG examples.
We include the following models as our baselines:

\begin{itemize}
\item \textbf{\xlm{}} Fine-tuning XLM with the English QG data.
\item \textbf{\txlmt{}} The pipeline of translating input Chinese sentences into English first, then performing En-En-QG with the XLM model, and finally translating back to the Chinese. We use the Transformer as the translator, which is also trained on the MultiUN dataset.
\item \textbf{\txlmt{} with Google Translator} Utilizing Google Translator in \txlmt{} for translation.
\end{itemize}



We evaluate models by both automatic evaluation metrics and human experts.
The automatic metrics scores are computed by regarding each Chinese character as a token.
For human evaluation, we consider three metrics: relatedness, fluency, and correctness, which are represented as integers ranged from 1 to 3. 
We randomly select 100 passage-answer pairs from the English QG test set, and use the models to generate questions.
Then we present these examples to three experts to ask for the above scores.
In Table~\ref{table:zh-zh-qg} and Table~\ref{table:human-zh}, we present the results for the zero-shot Zh-Zh-QG.
The results of monolingual supervised models are also reported in Table~\ref{table:en-en-qg} as reference.
In the automatic evaluation, our model consistently performs better than baselines in both zero-shot and monolingual supervised setting.
In the human evaluation, our model also obtains significant improvements in terms of relatedness and correctness.

\paragraph{English-Chinese Question Generation}
In the zero-shot English-Chinese question generation experiments, we use \xlm{} and \xlmt{} as our baselines. 
\xlmt{} is a pipeline method that uses En-En-QG with \xlm{} to generate questions, and then translates the results to Chinese.
Because there are no annotations for En-Zh-QG, we perform human evaluation studies for this setting.
Table~\ref{table:en-zh-qg} shows the human evaluation results, where our model surpasses all the baselines  especially in terms of relatedness and correctness.

\paragraph{Chinese-English Question Generation}
We also conduct experiments for zero-shot Chinese-English question generation, and adopt the same evaluation procedure to En-Zh-QG.
\txlm{} first translates Chinese input to English, and then conduct En-En-QG with \xlm{}.
As shown in Table~\ref{table:zh-en-qg}, human evaluation results indicate that \ours{} achieves significant improvements on the three metrics.


\subsection{Abstractive Summarization}

We conduct experiments on cross-lingual abstractive summarization (AS). AS is the task of converting the input sentences into summaries while preserving the key meanings.
For evaluation, we use English/French/Chinese Gigaword\footnote{LDC2011T07, LDC2011T10, LDC2011T13} to extract the first sentence and the headline of each article, and regard them as input document and predicted summaries, respectively.
For each language, we sample 500k/5k/5k examples for training/validation/test.

\paragraph{Zero-Shot Summarization}

In the zero-shot setting, we only use English data for training, and directly evaluate the model on other languages.
In Table~\ref{table:fr-fr-as} and Table~\ref{table:zh-zh-as}, we present the results for French/Chinese AS, which are evaluated by the ROUGE-1, ROUGE-2 and ROUGE-L metrics. 
We also report the results of supervised AS in Table~\ref{table:mono-as} for reference.
We find that \ours{} outperforms all the baseline models on both French and Chinese AS.
Comparing with French, there is a larger gap between baselines and our model on zero-shot Chinese AS, which indicates that the error propagation issue is more serious on distant language pairs.

\subsection{Ablation Studies}

\paragraph{Effects of Pre-Training}

\begin{table}[t]
\centering
\small
\begin{tabular}{lccc}
\toprule
Models     &  BL-4           & MTR            & RG-L        \\ \midrule
\xlm{}       &  0.25           & 0.62           & 2.56           \\ \midrule
\ours{}      &  \textbf{16.37} & \textbf{18.74} & \textbf{34.93} \\
~~~ $-$~XAE  &  13.71          & 15.88          & 31.43          \\
~~~ $-$~DAE  &  0.38           & 1.79           & 3.79           \\ \bottomrule
\end{tabular}
\caption{Ablations for pre-training objectives, where models are evaluated on zero-shot Chinese-Chinese question generation. Same shorthands apply as in Table~\ref{table:en-en-qg}.}
\label{table:ablation}
\end{table}

We conduct ablation studies for pre-training objectives, and the results can be seen in Table~\ref{table:ablation}.
We observe that our model greatly benefits from the DAE objective for the zero-shot Chinese question generation task.
The results also demonstrate that combining DAE and XAE can alleviate the spurious correlation issue and improves cross-lingual NLG.


\paragraph{Effects of Fine-Tuning Strategies}

\begin{table}[t]
\centering
\small
\begin{tabular}{lcccccc}
\toprule
                    & \multicolumn{3}{c}{Supervised En-En-QG}                     & \multicolumn{3}{c}{Zero-Shot Zh-Zh-QG}                     \\
                    & BL-4           & MTR            & RG-L           & BL-4           & MTR            & RG-L        \\ \midrule
All                & \textbf{19.99} & \textbf{24.05} & 48.74 & 6.82           & 14.84          & 21.77          \\
Dec             & 16.37          & 20.91          & 44.51          & 0.21           & 1.25           & 2.05           \\
Enc             & 19.62          & 23.66          & \textbf{48.78}          & 15.72          & \textbf{18.89} & 34.82          \\
ET & 19.69          & 23.73          & 48.53          & \textbf{16.37} & 18.74          & \textbf{34.93} \\ \bottomrule
\end{tabular}
\caption{Effects of different fine-tuning strategies. Dec, Enc and ET represent fine-tuning the parameters of the decoder, the encoder, and the Transformer layers of the encoder, respectively. Same shorthands apply as in Table~\ref{table:en-en-qg}.}
\label{table:ft}
\end{table}

As shown in Table~\ref{table:ft}, we use the En-En-QG and Zh-Zh-QG tasks to analyze the effects of using different fine-tuning strategies.
It can be observed that fine-tuning encoder parameters, our model obtain an impressive performance for both English and Chinese QG, which shows the strong cross-lingual transfer ability of our model.
When fine-tuning all the parameters, the model achieves the best score for English QG, but it suffers a performance drop when evaluating on Chinese QG.
We find that fine-tuning decoder hurts cross-lingual decoding, and the model learns to only decode English words.
For only fine-tuning decoder, the performance degrades by a large margin for both languages because of the underfitting issue, which indicates the necessity of fine-tuning encoder.

\paragraph{Effects of Cross-Lingual Transfer}

\begin{figure}[t]
\begin{center} 
\includegraphics[width=0.96\linewidth]{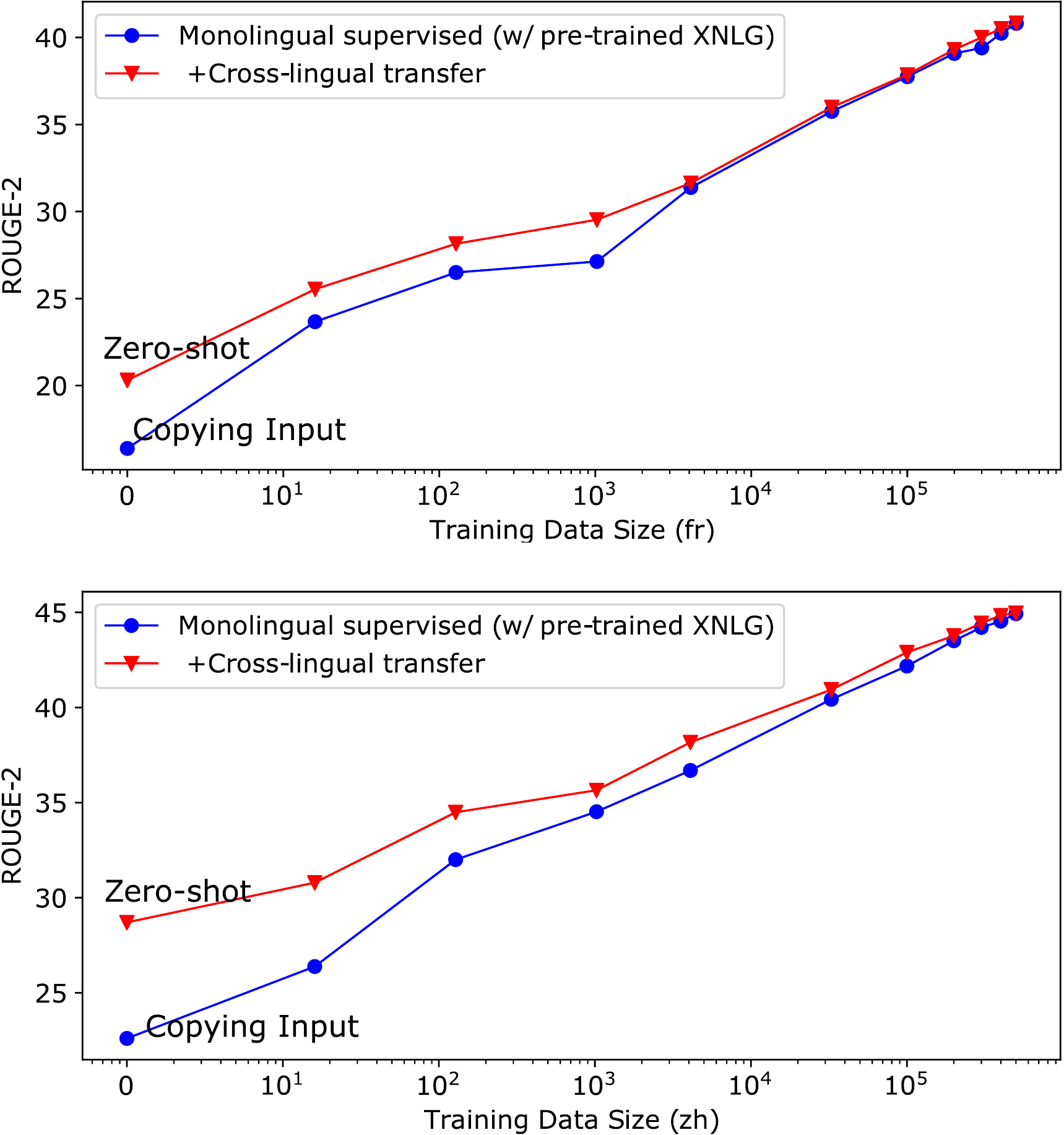}
\caption{ROUGE-2 scores for few-shot French/Chinese abstractive summarization with different training data sizes.
} 
\label{fig:semi}
\end{center} 
\end{figure}

\begin{figure*}[th]
\begin{center} 
\includegraphics[width=1\linewidth]{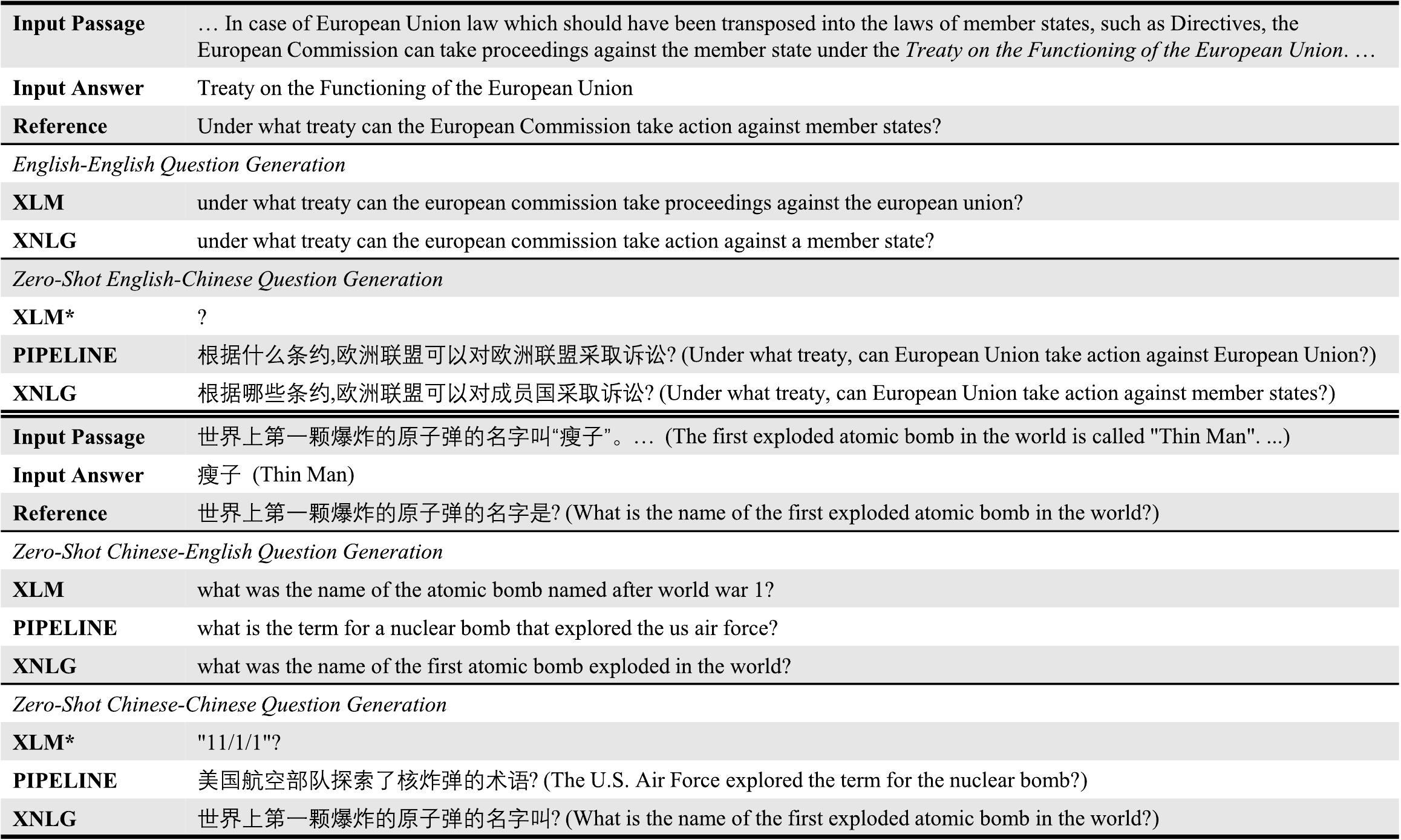}
\caption{Examples of generated questions by \ours{} and the baselines in four directions (En-En,En-Zh,Zh-En and Zh-Zh). ``*'': Because \xlm{} is not designed for cross-lingual NLG, it is hard to produce meaningful sentences for En-Zh-QG and Zh-Zh-QG.
} 
\label{fig:case}
\end{center} 
\end{figure*}

We examine whether low-resource NLG can benefit from cross-lingual transfer.
We consider English as the rich-resource language, and conduct experiments for few-shot French/Chinese AS.
Specifically, we first fine-tune \ours{} on the English AS data, and then fine-tune it on the French or Chinese AS data.
We compare with the monolingual supervised model that \ours{} is only fine-tuned on the dataset of the target language.
As shown in Figure~\ref{fig:semi}, we can observe that the cross-lingual supervision improves performance for few-shot abstractive summarization.
As the training data size becomes larger, the performances of the two models are getting closer.

\subsection{Case Studies}

As shown in Figure~\ref{fig:case}, we present some examples generated by \ours{} and the baselines in four directions (En-En, En-Zh, Zh-En, and Zh-Zh). 
When decoding on an unseen language, \xlm{} tends to generate random output, because it is not designed for cross-lingual NLG. 
In terms of the pipeline model, we can observe that it suffers from the error propagation issue, especially when the source and target languages are all different from the training data. 
For example, when the pipeline model performs Zh-Zh-QG, keywords are translated twice, increasing the risk of mistranslation. 
In the second example, ``\textit{atomic bomb}'' is mistranslated to ``\textit{nuclear bomb}'', resulting in its low correctness. 
On the contrary, by directly transferring English supervision signals to the other generation directions, the generated questions of \ours{} match the references better than baselines.

\section{Conclusion}
In this paper, we propose a pre-training method for cross-lingual natural language generation (NLG) that can transfer monolingual NLG supervision signals to all pre-trained languages.
With the pre-trained model, we achieve zero-shot cross-lingual NLG on several languages by only fine-tuning once.
Experimental results show that our model outperforms the machine-translation-based pipeline model on several cross-lingual NLG tasks.
For future work, we would like to improve our pre-training method towards the fully unsupervised setting.

\section*{Acknowledgements}
Prof. Heyan Huang is the corresponding author. The work is supported by NKRD (No. 2018YFB1005100), NSFC (No. 61772076 and 61751201), NSFB (No. Z181100008918002), Major Project of Zhijiang Lab (No. 2019DH0ZX01), and Open fund of BDAlGGCNEL and CETC Big Data Research Institute Co., Ltd  (No. w-2018018).

\bibliographystyle{aaai}
\bibliography{xnlg}

\end{document}